\documentclass[11pt]{article}
\usepackage[preprint]{acl}

\usepackage{times}
\usepackage{latexsym}
\usepackage[T1]{fontenc}
\usepackage[utf8]{inputenc}
\usepackage{microtype}
\usepackage{graphicx}
\usepackage{booktabs}
\usepackage{multirow}
\usepackage{amsmath}
\usepackage{amssymb}
\usepackage{enumitem}
\usepackage{algorithm}
\usepackage{algorithmic}
\usepackage{tikz}
\usepackage{xcolor}
\usepackage{url}
\usetikzlibrary{arrows.meta, positioning, fit, calc,
                shapes.geometric, backgrounds, decorations.pathreplacing}

\newcommand{\method}{HindSearch}
\newcommand{\TLHC}{TLHC}

\title{\method{}: Trajectory-Level Hindsight Critique for\\Search-Augmented Reinforcement Learning}

\author{Haowei Liu \\
  Santa Clara University \\
  Santa Clara, CA \\
  hliu6@scu.edu \\\And
  Jiamian Wang \\
  Rochester Institute of Technology \\
  Rochester, NY \\
  jw4905@rit.edu \AND
  Hsin-Tai Wu \\
  Independent Researcher \\
  Sunnyvale, CA \\
  htwunew@gmail.com \\\AND
  Zhiqiang Tao \\
  Rochester Institute of Technology \\
  Rochester, NY \\
  zhiqiang.tao@rit.edu \\\And
    Yi Fang \\
  Santa Clara University \\
  Santa Clara, CA \\
  \texttt{yfang@scu.edu} \\
  }

\begin{document}
\maketitle

\begin{abstract}
Search-augmented LM agents are typically trained with a binary exact-match reward, which signals only whether a trajectory succeeded without telling us why it failed.
We introduce \method{}, a hindsight self-distillation procedure for GRPO: after each rollout, a frozen judge writes a short critique of every failed trajectory using the gold answer, and the critique supplies an auxiliary on-policy distillation signal on the student's search actions.
On the standard seven-benchmark suite with Qwen2.5-3B-Instruct, \method{} reaches \textbf{39.4}\% average EM, outperforming prior search-RL baselines. Removing the judge's access to the gold answer erases most of the gain, isolating hindsight as the source of the improvement. Code is available at \url{https://anonymous.4open.science/r/hindsearch-anon-EBDC}.
\end{abstract}

\section{Introduction}
\label{sec:intro}

Search-augmented LM agents interleave retrieval with reasoning and are commonly trained end-to-end under a scalar outcome reward \citep{jin2025searchr1, rsearch2025, autorefine2025, asearcher2026, zerosearch2025}.
The setup is simple but throws away signal. A failed trajectory contains the queries the agent issued, the snippets they returned, and—at training time—the gold answer; together these often pin down what should have been searched for instead. An outcome-only objective ignores all of this and asks the agent to recover the underlying search policy from a bandit signal, which matches the long credit-assignment delay and the stability issues observed in publicly released search-RL training runs.

\begin{figure}[t]
\centering
\scriptsize
\definecolor{agentC}{HTML}{5B6B7A}  
\definecolor{teachC}{HTML}{B8843A}  
\definecolor{hintC}{HTML}{3E7C7A}   

\begin{tikzpicture}[
  font=\scriptsize,
  >={Latex[length=1.4mm]},
  every node/.style={inner sep=2pt},
  box/.style    ={rounded corners=1.5pt, draw, line width=0.4pt,
                  align=left, inner sep=3pt},
  qbox/.style   ={box, fill=agentC!8,  draw=agentC!70},
  jbox/.style   ={box, fill=teachC!18, draw=teachC!90, line width=0.7pt},
  cbox/.style   ={box, fill=teachC!10, draw=teachC!75, align=center},
  tbox/.style   ={box, fill=hintC!10,  draw=hintC!75,  align=center},
  sbox/.style   ={box, fill=agentC!12, draw=agentC!75, align=center},
  gtbox/.style  ={box, fill=hintC!12,  draw=hintC!75,  align=center},
  arr/.style    ={->, draw=agentC!75, line width=0.4pt},
  harr/.style   ={->, draw=teachC!90, line width=0.55pt},
  karr/.style   ={->, draw=hintC!95,  line width=0.6pt, dashed},
  lbl/.style    ={font=\tiny\itshape, text=black!70},
  klbl/.style   ={font=\tiny\itshape, text=hintC!60!black},
  hlbl/.style   ={font=\tiny\itshape, text=teachC!50!black}
]

\node[qbox, text width=0.94\linewidth] (panelA) {%
  \textbf{\textcolor{agentC!85!black}{Failed rollout}}
  \hfill \textcolor{agentC!75!black}{scalar reward $r{=}0$ (EM)}\\[1pt]
  \textbf{Q:} Who directed the Best Picture winner at the 80th Academy Awards?\\[1pt]
  {\color{agentC!85!black}\texttt{[s$_1$]}} \texttt{search}: ``Best Picture 80th Academy Awards''\\
  \hspace*{2.6mm}$\rightarrow$ \emph{No Country for Old Men (2007)}\\
  {\color{agentC!85!black}\texttt{[s$_2$]}} \texttt{search}: ``No Country for Old Men \underline{cast}''\\
  \hspace*{2.6mm}$\rightarrow$ \emph{Javier Bardem, Tommy Lee Jones, \ldots}\\
  \texttt{ans:} \textcolor{red!55!black}{\texttt{<answer>\,Javier Bardem\,</answer>}}
  \quad\textcolor{red!55!black}{\textbf{\texttt{X}}}
};

\node[gtbox, below=4mm of panelA.south west, anchor=north west,
      text width=0.28\linewidth]
  (gt) {\textbf{Gold} $a^\star$\\ \emph{Coen Brothers}};

\node[jbox, right=3mm of gt, text width=0.32\linewidth, align=center,
      minimum height=8mm]
  (judge) {\textbf{\textcolor{teachC!60!black}{Judge LM}}\\[-1pt]
           \tiny hindsight critic};

\node[box, fill=agentC!8, draw=agentC!60,
      right=3mm of judge, text width=0.18\linewidth, align=center]
  (tauchip) {trajectory\\ $\tau$};

\node[cbox, below=3.2mm of judge, text width=0.94\linewidth]
  (crit) {\textbf{\textcolor{teachC!55!black}{Critique $h$:}}
  \emph{``After identifying the film, query the director directly
  (`No Country for Old Men director'), not the cast.''}};

\node[tbox, below=3.5mm of crit.south west, anchor=north west,
      text width=0.44\linewidth]
  (teach) {\textbf{Frozen Teacher}\\ $\pi_T(\cdot \mid \tau_{<t},\,h)$\\[1pt]
  \tiny up-weights \texttt{``director''}, \texttt{``Coen''}};

\node[sbox, below=3.5mm of crit.south east, anchor=north east,
      text width=0.40\linewidth, minimum height=10mm]
  (stud) {\textbf{Student} $\pi_\theta$\\[1pt]
  \tiny search-token loss only};

\draw[arr] (panelA.south) ++(0,-0.6mm) -- ++(0,-1mm)
   -| (tauchip.north);
\draw[harr] (tauchip.west) -- (judge.east);

\draw[harr] (gt.east) -- (judge.west);

\draw[harr] (judge.south) -- (crit.north)
   node[hlbl, midway, right] {hindsight};

\draw[harr] (crit.south -| teach.north) -- (teach.north)
   node[hlbl, midway, right] {condition};

\draw[karr] (teach.east) -- (stud.west)
   node[klbl, midway, above] {$D_{\mathrm{KL}}$ (OPD)};

\end{tikzpicture}

\caption{\textbf{\method{} at a glance.}
Standard GRPO collapses a failed multi-turn rollout (top) into a single $r{=}0$ scalar, discarding \emph{how} the agent erred.
\method{} feeds the full trajectory $\tau$ and the gold answer $a^\star$ to a frozen judge, which emits a directive critique $h$.
The critique conditions a frozen self-teacher whose action distribution is distilled into the student via on-policy KL (OPD), masked to search-action tokens, recovering rich corrective signal from rollouts that would otherwise produce no reward gradient.}
\label{fig:hindsearch}
\end{figure}
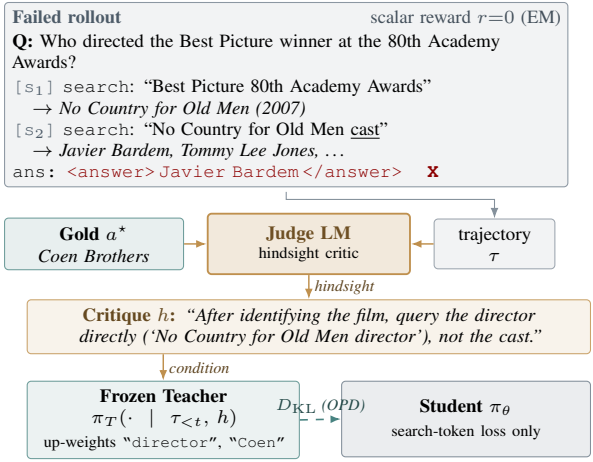

Prior work on denser signals divides roughly into process reward models for retrieval \citep{prorag2026, searchp1_2026, smartsearch2026, evalact2026}, which assign scalar quality scores to individual steps, and hint-conditioned distillation methods such as OpenClaw-RL \citep{openclawrl2026}, which translates a process reward model's natural-language hint into token-level supervision. Process rewards remain evaluative—they grade a step rather than say what to do instead—and OpenClaw-RL's hints are produced online from the next state, without the gold answer. It is also validated on GUI control and tool-call domains, where each step has a clear local notion of success. None of this exploits the cleanest signal we have at training time for search: the answer itself.

We propose \textbf{trajectory-level hindsight critique (\TLHC{})}: after each GRPO rollout, every failed trajectory is sent to a frozen judge together with its gold answer, and the judge returns a short critique of what the agent should have done differently. The critique conditions a frozen teacher whose log-probabilities supply an auxiliary on-policy distillation (OPD) \citep{agarwal2024onpolicydistillationlanguagemodels} loss on the student's search-action tokens. Because the judge sees the answer, the critique is directive rather than evaluative; because it is produced post hoc, no online process reward model is needed. Section~\ref{sec:method} gives the full procedure.

Our contributions are:
\begin{itemize}[leftmargin=*,itemsep=0pt,topsep=2pt]
\item \textbf{TLHC:} a post-hoc hindsight-critique procedure that converts trajectory-level EM failures into per-token OPD supervision on top of GRPO.
\item \textbf{Empirical:} \textbf{39.4\%} average EM on the standard seven-benchmark search-QA suite, outperforming prior search-RL baselines, with stable training throughout the $300$-step budget.
\end{itemize}

\section{Related Work}
\label{sec:related}

\paragraph{RL for search agents and process rewards.}
Search-R1 \citep{jin2025searchr1} introduced the multi-turn search-RL recipe — PPO \citep{ppo} or GRPO \citep{grpo} under a binary EM reward — that subsequent work adopts as a common starting point. Follow-up work explores variants of this recipe: outcome-only setups \citep{r1searcher2025, research2025}, multi-reward shaping \citep{rsearch2025}, refinement steps \citep{autorefine2025}, longer horizons \citep{asearcher2026, agentgym_rl_2025}, decoupling searcher from generator \citep{s3_2025}, knowledge-boundary-aware gating \citep{ikea2025}, simulated retrieval for cheaper training \citep{zerosearch2025}, live-web environments \citep{deepresearcher2025}, and parallel multimodal search \citep{li2026hypereyesdualgrainedefficiencyawarereinforcement}. A separate line introduces step-level process rewards: information-gain shaping \citep{stepsearch2025, igsearch2026}, turn-level credit assignment \citep{tspo2026}, and other learned step scorers \citep{prorag2026, searchp1_2026, smartsearch2026, evalact2026, cui2025prime}. These signals are denser than EM but still evaluative—they grade a step rather than instruct a correction.

\paragraph{Hindsight and hint-conditioned distillation.}
Hindsight experience replay \citep{her1997} relabels trajectories with achieved goals; AgentHER \citep{agenther2026} and ECHO \citep{echo2025} adapt the idea to LM agents via off-policy rewriting. Concurrent work on self-distillation for search agents includes SD-Search \citep{ma2026sdsearchonpolicyhindsightselfdistillation} and Search-E1 \citep{liang2026searche1selfdistillationdrivesselfevolution}. The closest prior method is OpenClaw-RL \citep{openclawrl2026}, which conditions a frozen same-family teacher on a natural-language hint and distills the resulting log-probability gap into the student. We adopt its hint-prefix teacher mechanism but differ in two ways. Beyond the post-hoc-vs-online hint source already noted in \S\ref{sec:intro}, OpenClaw-RL folds the directive signal into a PPO-clipped surrogate restricted to a top-$K\!=\!4$ vocabulary subset, whereas we use a separate, plain-clamp OPD loss on the full response distribution, gated to failed trajectories and masked to search-action tokens. We ablate the loss form in \S\ref{sec:results}.

\section{Method}
\label{sec:method}

\subsection{Problem setup}
We study multi-turn search-augmented QA with a single LM policy $\pi_\theta$.
At each turn $t$, the policy emits either a search action $\langle\textsc{search}\rangle q_t \langle/\textsc{search}\rangle$, whose retrieved result $o_t$ is appended to the context as an $\langle\textsc{information}\rangle$ block, or a final answer $\langle\textsc{answer}\rangle a\langle/\textsc{answer}\rangle$.
A trajectory $\tau=(q_1,o_1,\ldots,a)$ receives a binary reward $r(\tau)=\mathbb{I}[\mathrm{EM}(a,a^\star)]$ against gold answer $a^\star$.

\subsection{GRPO with search rollouts}
We optimize $\pi_\theta$ with GRPO \citep{grpo} under the standard search-RL recipe \citep{jin2025searchr1}.
For each prompt, $G$ trajectories are sampled, group-normalized advantages are computed from outcome rewards, and retrieved $\langle\textsc{information}\rangle$ tokens are masked from the policy gradient.
Let $L_{\mathrm{PPO}}(\theta)$ denote the standard clipped policy-gradient objective.
\method{} adds a single auxiliary term on top.

\begin{table*}[t]
\centering
\small
\setlength{\tabcolsep}{4pt}
\begin{tabular}{lccccccc c}
\toprule
Method & NQ$^\dagger$ & TriviaQA & PopQA & HotpotQA$^\dagger$ & 2Wiki & MuSiQue & Bamboogle & Avg \\
\midrule
Direct Inference & 10.6 & 28.8 & 10.8 & 14.9 & 24.4 & \phantom{0}2.0 & \phantom{0}2.4 & 13.4 \\
CoT~\citep{wei2022cot} & \phantom{0}2.3 & \phantom{0}3.2 & \phantom{0}0.5 & \phantom{0}2.1 & \phantom{0}2.1 & \phantom{0}0.2 & \phantom{0}0.0 & \phantom{0}1.5 \\
IRCoT~\citep{trivedi2023ircot} & 11.1 & 31.2 & 20.0 & 16.4 & 17.1 & \phantom{0}6.7 & 24.0 & 18.1 \\
Search-o1~\citep{li2025searcho1} & 23.8 & 47.2 & 26.2 & 22.1 & 21.8 & \phantom{0}5.4 & $\mathbf{32.0}$ & 25.5 \\
RAG~\citep{lewis2020rag} & 34.8 & 54.4 & 38.7 & 25.5 & 22.6 & \phantom{0}4.7 & \phantom{0}8.0 & 27.0 \\
SFT & 24.9 & 29.2 & 10.4 & 18.6 & 24.8 & \phantom{0}4.4 & 11.2 & 17.6 \\
R1-instruct~\citep{guo2025deepseekr1} & 21.0 & 44.9 & 17.1 & 20.8 & 27.5 & \phantom{0}6.0 & 19.2 & 22.4 \\
Rejection Sampling & 29.4 & 48.8 & 33.2 & 24.0 & 23.3 & \phantom{0}5.9 & 21.0 & 26.5 \\
Search-R1-Instruct~\citep{jin2025searchr1} & 39.7 & 56.5 & 39.1 & 33.1 & 31.0 & 12.4 & 23.2 & 33.6 \\
Search-R1-Base~\citep{jin2025searchr1} & 42.1 & 58.3 & 41.3 & 29.7 & 27.4 & \phantom{0}6.6 & 12.8 & 31.2 \\
ReSearch-Instruct~\citep{research2025} & 36.5 & 57.1 & 39.5 & 35.1 & 27.2 & \phantom{0}9.5 & 26.6 & 33.1 \\
ReSearch-Base~\citep{research2025} & 42.7 & 59.7 & 43.0 & 30.5 & 27.2 & \phantom{0}7.4 & 12.8 & 31.9 \\
\textbf{\method{} } & $\mathbf{45.0}$ & $\mathbf{60.5}$ & $\mathbf{45.2}$ & $\mathbf{39.5}$ & $\mathbf{40.1}$ & $\mathbf{14.2}$ & 31.2 & $\mathbf{39.4}$ \\
\bottomrule
\end{tabular}
\caption{Performance comparison across seven QA benchmarks (Qwen2.5-3B backbone, Instruct unless suffixed -Base). Best results in each column are in \textbf{bold}. $^\dagger$ in-domain training data. All methods share the same retriever (E5/Wiki-18), top-$k\!=\!3$, training corpus (NQ+HotpotQA), and strict-EM scorer.}
\label{tab:main}
\end{table*}

\subsection{Trajectory-level hindsight critique}
\label{sec:tlhc}

After each rollout batch, we partition trajectories into $\mathcal{T}_+=\{\tau:r(\tau)=1\}$ and $\mathcal{T}_-=\{\tau:r(\tau)=0\}$.
For each $\tau\in\mathcal{T}_-$, a frozen judge $\pi_J$ is queried with the rendered trajectory and the gold answer to produce a 1--2 sentence critique:
\begin{equation}
h(\tau) \;=\; \pi_J\!\big(\tau,\,a^\star\big).
\label{eq:judge}
\end{equation}
The teacher $\pi_{\mathrm{ref}}$ is the student's frozen initialization (also reused as the GRPO KL anchor).
For every search step $t$ in $\tau$, the hint is broadcast as a prefix:
\begin{equation}
c^T_t \;=\; \langle\textsc{hint}\rangle\, h(\tau)\, \langle/\textsc{hint}\rangle \,\oplus\, c_t.
\label{eq:tctx}
\end{equation}
With $\ell^s_t=\log\pi_\theta(a_t\mid c_t)$ and $\ell^T_t=\log\pi_{\mathrm{ref}}(a_t\mid c^T_t)$, the OPD loss is
\begin{equation}
L_{\mathrm{OPD}} \;=\; -\frac{1}{|M|}\!\sum_{t\in M}\!\mathrm{clamp}\!\big(\ell^T_t-\ell^s_t,\,0,\,1\big)\cdot\ell^s_t,
\label{eq:opd}
\end{equation}
where $M$ is the set of response tokens inside a $\langle\textsc{search}\rangle\!\ldots\!\langle/\textsc{search}\rangle$ span within a failed trajectory. The $\mathrm{clamp}(\cdot)$ coefficient is computed with a stop-gradient on $\ell^T_t-\ell^s_t$, so the only differentiable factor is $\ell^s_t$; the update therefore raises $\ell^s_t$ on tokens where the teacher is more confident than the student.
The lower clamp restricts gradient flow to tokens where the hint-conditioned teacher is more confident than the student; the upper clamp prevents outliers from dominating the batch.
Answer tokens are left to GRPO.
The total objective is
\begin{equation}
L_{\mathrm{total}} \;=\; L_{\mathrm{PPO}} \;+\; \lambda_{\mathrm{OPD}}\cdot L_{\mathrm{OPD}},
\label{eq:total}
\end{equation}
with $\lambda_{\mathrm{OPD}}=0.01$ in all main runs.
Appendix~\ref{app:algo} gives the full algorithm.

\section{Experiments}
\label{sec:exp}

\paragraph{Datasets.}
We evaluate \method{} on seven question-answering benchmarks covering both single-hop and multi-hop settings: NQ \citep{kwiatkowski2019nq}, TriviaQA \citep{joshi2017triviaqa}, PopQA \citep{mallen2023popqa}, HotpotQA \citep{yang2018hotpotqa}, 2WikiMultihopQA \citep{ho2020twowiki}, MuSiQue \citep{trivedi2022musique}, and Bamboogle \citep{press2022bamboogle}.
Training data are the merged NQ and HotpotQA training splits (169k samples); the other five benchmarks are zero-shot.
We follow prior work and adopt exact match (EM) as the evaluation metric, with $51{,}713$ validation samples in total, validated every $50$ training steps.

\paragraph{Baselines.}
We compare against a full set of baselines spanning (1) inference without retrieval (Direct, CoT \citep{wei2022cot}); (2) inference with retrieval (RAG \citep{lewis2020rag}, IRCoT \citep{trivedi2023ircot}, Search-o1 \citep{li2025searcho1}); (3) fine-tuning based methods (SFT, R1-style RL without search \citep{guo2025deepseekr1}, rejection sampling, and the Search-R1 PPO and GRPO variants \citep{jin2025searchr1}); and (4) the concurrent simulated-retrieval method ZeroSearch \citep{zerosearch2025}, whose released checkpoint we evaluate under our retriever.
To ensure a fair comparison, all methods use the same backbone, retriever, knowledge corpus, top-$k$, and training data.

\paragraph{Training details.}
The backbone is Qwen2.5-3B-Instruct, with E5-base-v2 \citep{wang2022e5} as the retriever over the Wikipedia-18 corpus, top-3 retrieval, and a maximum of $T\!\le\!4$ turns per trajectory.
We use $\pi_\theta=\pi_{\mathrm{ref}}=$ Qwen2.5-3B-Instruct \citep{qwen2.5} and Qwen3.6-27B \citep{qwen3_6} as the judge $\pi_J$, served via a local API.
$\lambda_{\mathrm{OPD}}=0.01$, GRPO group size $G=8$, and we train for $300$ steps on $2\!\times\!$H200.
The OPD pass adds roughly $15\%$ wall-clock per step.
Full hyperparameters in Appendix~\ref{app:hparams}.

\section{Results}
\label{sec:results}

\begin{figure}[t]
\centering
\includegraphics[width=0.95\linewidth]{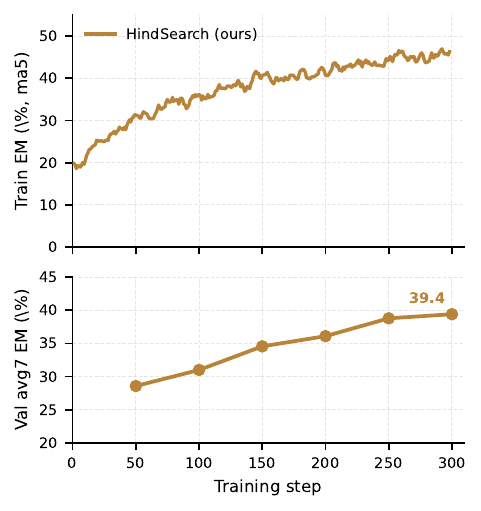}
\caption{\method{} training and validation dynamics across $300$ GRPO steps. Top: train EM (rolling 5-step mean) climbs steadily to a peak of $49.97$. Bottom: val avg EM (dots) rises monotonically from $28.6$ at step $50$ to $39.4$ at step $300$.}
\label{fig:curve}
\end{figure}

\paragraph{Main result.}
Table~\ref{tab:main} reports validation EM averaged across the seven benchmarks.
\method{} exceeds all prior baselines by step~$150$ and continues climbing throughout training (Figure~\ref{fig:curve}).
\method{} reaches $39.4$ at its best checkpoint, ahead of the strongest prior baseline (Search-R1 GRPO, $33.6$) by $+5.8$ points.
Training was stable through step $300$.

\paragraph{Training curve.}
Figure~\ref{fig:curve} shows \method{}'s training and validation dynamics over the $300$-step run. Train EM (rolling 5-step mean) climbs steadily to a peak of $49.97$, with KL to the anchor staying below $0.43$ throughout. Val avg EM rises monotonically from $28.6$ at step $50$ to $39.4$ at step $300$, with no plateau visible at the end of training. The OPD term yields a smooth, stable curve under the same outcome reward and GRPO trust region.

\begin{table}[t]
\centering
\small
\setlength{\tabcolsep}{4pt}
\begin{tabular}{lc}
\toprule
Configuration & Avg EM \\
\midrule
\textbf{\method{} (full, 3B-Instruct)} & $\mathbf{39.4}$ \\
\midrule
$-$ hindsight (judge w/o $a^\star$) & \texttt{34.7} \\
PPO-clipped OPD \citep{openclawrl2026} & \texttt{32.7} \\
Multi-seed (seed $42$) & \texttt{39.2} \\
\midrule
Per-step earliest-critical & $35.5$ \\
7B teacher & collapse \\
\bottomrule
\end{tabular}
\caption{Ablations on the seven-benchmark search-QA suite (val@$300$). \emph{7B teacher} collapsed within tens of steps. Additional variants in Appendix~\ref{app:negative}.}
\label{tab:ablation}
\end{table}

\paragraph{Per-task gains.}
Gains show up on most benchmarks: TriviaQA leads at $60.5$, with five others above $30$. MuSiQue, the hardest multi-hop benchmark, reaches $14.2$—low in absolute terms but in line with prior reports on its difficulty. The improvement transfers cleanly off the training distribution: the two in-domain benchmarks (NQ, HotpotQA) gain $+5.85$ on average over Search-R1 GRPO, while the five zero-shot benchmarks gain $+5.80$, indicating that the OPD signal induces general search behavior rather than memorization of training-set surface patterns.

\paragraph{Ablations.}
Table~\ref{tab:ablation} isolates four design choices.
\emph{Hindsight.} Removing the judge's gold-answer access drops the average from $39.4$ to $34.7$, erasing $4.7$ of the $5.8$-point gain over Search-R1 GRPO. This is consistent with hindsight, rather than the act of judging alone, being the primary contributor (a no-answer judge also produces less directive critiques, which is partially confounded with the information change).
\emph{Loss form.} Swapping the plain-clamp OPD loss for OpenClaw-RL's PPO-clipped, top-$K$ surrogate \citep{openclawrl2026} drops the average to $32.7$, below Search-R1 GRPO: the loss form is load-bearing. We caution that this re-implements an OPD loss designed for GUI/tool-use into search-QA, so the $-6.7$pp gap may include a loss-form $\times$ task-domain interaction.
\emph{Broadcast vs.\ per-step hint.} A per-step ``earliest critical'' variant reaches $35.5$, $3.9$ points below the full method; broadcasting the critique to every search step is a non-trivial design choice.
\emph{Teacher choice.} A same-family 7B teacher destabilizes training within tens of steps (Appendix~\ref{app:negative}).

\paragraph{Sensitivity analyses.}
\emph{Judge size and family.} Replacing Qwen3.6-27B \citep{qwen3_6} with same-family Qwen2.5-7B-Instruct \citep{qwen2.5} drops the average to $34.8$ ($-4.6$ pp); a cross-family DeepSeek-V4-Flash \citep{deepseek_v4} judge reaches $39.6$ ($+0.2$ pp). The method transfers across judge families but degrades when the judge is substantially weaker than the student's critique-following capacity.
\emph{Answer leakage.} The judge sees the gold answer, so its critique can surface the answer string. Three facts bound the concern: the judge is absent at inference; the OPD loss is masked to search-action tokens and never updates the final-answer span; and the in-domain ($+5.85$pp) and zero-shot ($+5.80$pp) gains are nearly identical, inconsistent with memorization-via-leakage as the dominant mechanism.

\section*{Limitations}
\label{sec:limitations}

\paragraph{Backbone and scale.}
All main experiments use Qwen2.5-3B-Instruct; Table~\ref{tab:ablation} includes a second-seed run. We focus on the instruction-tuned variant: applying \method{} from a pure base model (Qwen2.5-3B-Base) requires a format-acquisition stage---GRPO and the OPD signal can only act once the policy emits well-formed \texttt{<search>}/\texttt{<answer>} actions---which is out of scope for this short paper.

\paragraph{Retriever and concurrent baselines.}
We use E5-base-v2 \citep{wang2022e5} over the Wikipedia-18 corpus, top-$k\!=\!3$, following the standard search-RL evaluation setup \citep{jin2025searchr1, research2025, autorefine2025}. Numbers may shift under sparse retrievers (e.g., BM25). Concurrent work such as ZeroSearch \citep{zerosearch2025} targets an orthogonal axis---training-time retrieval cost---under a different retrieval setup; a head-to-head comparison would require matching its retriever and reward, and we do not attempt it here.

\bibliography{references}

\newpage
\appendix

\section{Algorithm pseudocode (extended)}
\label{app:algo}

We give the full \method{} training step, with bookkeeping omitted from the main-paper sketch.

\begin{algorithm}[h]
\small
\caption{\method{} training step (full)}
\label{alg:hindsearch}
\begin{algorithmic}[1]
\REQUIRE policy $\pi_\theta$, frozen teacher $\pi_{\mathrm{ref}}$, judge $\pi_J$, batch $\mathcal{B}$, group size $G$, weight $\lambda_{\mathrm{OPD}}$
\STATE For each prompt $p\in\mathcal{B}$, rollout $G$ trajectories $\{\tau^{(p)}_g\}_{g=1}^G$ with $\pi_\theta$.
\STATE Compute EM rewards $r(\tau^{(p)}_g)$ and group-normalized advantages $\hat{A}^{(p)}_g$.
\STATE $\mathcal{T}_-\leftarrow\{\tau:r(\tau)=0\}$.
\FOR{$\tau\in\mathcal{T}_-$ in parallel}
  \STATE $h(\tau)\leftarrow\pi_J(\tau,a^\star_\tau)$ \COMMENT{single sample, temperature 0.7, max 96 tokens}
\ENDFOR
\STATE Build $M\leftarrow\{(\tau,t):\tau\in\mathcal{T}_-,\,a_t\in\langle\textsc{search}\rangle\text{ span}\}$.
\STATE For each $(\tau,t)\in M$, build $c^T_t$ via Eq.~\ref{eq:tctx}.
\STATE Forward $\pi_{\mathrm{ref}}$ on $\{c^T_t\}_{(\tau,t)\in M}$; collect $\ell^T_t$.
\STATE Forward $\pi_\theta$ on $\mathcal{B}$; collect $\ell^s_t$ on all response tokens.
\STATE $L_{\mathrm{PPO}}\leftarrow$ standard GRPO clipped objective with retrieved-token mask.
\STATE $L_{\mathrm{OPD}}\leftarrow$ Eq.~\ref{eq:opd} on $M$.
\STATE Step optimizer on $L_{\mathrm{PPO}} + \lambda_{\mathrm{OPD}} L_{\mathrm{OPD}}$.
\end{algorithmic}
\end{algorithm}

\section{Hyperparameters}
\label{app:hparams}

Table~\ref{tab:hparams} gives the full hyperparameter set used in all main and ablation runs.

\begin{table}[h]
\centering
\small
\begin{tabular}{ll}
\toprule
Setting & Value \\
\midrule
Backbone & Qwen2.5-3B-Instruct \\
Frozen teacher & Qwen2.5-3B-Instruct (init) \\
Judge model & Qwen3.6-27B \citep{qwen3_6} \\
Training data & NQ + HotpotQA (169k) \\
Retriever & E5-base-v2 (Wikipedia-18) \\
Top-$k$ retrieval & $3$ \\
Max turns $T$ & $4$ \\
RL algorithm & GRPO \\
Group size $G$ & $8$ \\
PPO clip $\epsilon$ & $0.2$ \\
KL coef (anchor) & $0.001$ \\
$\lambda_{\mathrm{OPD}}$ & $0.01$ \\
OPD token mask & search-span only, $\mathcal{T}_-$ only \\
OPD clamp range & $[0,1]$ \\
Optimizer & AdamW \\
LR & $1\!\times\!10^{-6}$ \\
LR schedule & constant \\
Train steps & $300$ \\
Validation every & $50$ steps \\
Hardware & $2\!\times\!$H200 \\
Wall-clock overhead & $\sim\!15\%$ vs.\ unmodified GRPO \\
\bottomrule
\end{tabular}
\caption{Full hyperparameters for the main \method{} runs.}
\label{tab:hparams}
\end{table}

\section{Negative results}
\label{app:negative}

We collect here variants we tried that did not work or underperformed the default \method{} configuration.

\paragraph{Per-step earliest-critical critique.}
Instead of broadcasting one trajectory-level critique $h(\tau)$ to every search step, we asked the judge to pick a single ``earliest critical'' search step and emitted a localized hint only there.
This variant reached $\texttt{val@50}=28.55$, comparable to the trajectory-broadcast value at the same step ($28.56$), but it never closed the gap and ended well below the trajectory-broadcast $\texttt{val@150}=34.55$.
Looking at intermediate checkpoints, the localized hint seems too narrow: when the critique is genuinely trajectory-level (e.g., ``the agent should have re-decomposed the question after the third turn''), pinning it to one token span fits poorly and the OPD gradient gets noisy.

\paragraph{Same-family 7B teacher.}
The obvious way to widen the teacher--student capability gap is a strictly larger same-family teacher. We tried Qwen2.5-7B-Instruct as $\pi_{\mathrm{ref}}$ with the 3B student. Training collapsed within \texttt{79} steps: validation EM dropped and KL to the anchor diverged.
Our reading is that the 7B teacher disagrees with the 3B student on routine search tokens regardless of the hint, and at $\lambda_{\mathrm{OPD}}=0.01$ this produces a persistent capability-gap gradient that pulls the student off the GRPO trust region rather than nudging it toward repair behaviors. The frozen 3B copy avoids this because its disagreement with the student is mediated entirely by the hint context.

\paragraph{Token-level judging.}
A variant differing from \method{} only in the judge output format—per-token categorical labels over the emitted search query instead of a natural-language critique—collapsed within tens of steps. Logs showed the per-token labels were dominated by a single class, producing a degenerate OPD target. We kept natural-language critiques in the main method.

\paragraph{Advantage-injection OPD on EM-rewarded search.}
We tried a direct port of OpenClaw-RL's advantage-injection formulation \citep{openclawrl2026} to our search setup, sweeping the injection weight and using both rule-based and judge-based hints. All injection variants collapsed within $25$--$85$ steps.
Inspecting the runs, on EM-rewarded search the outcome advantage and the directive log-probability gap often disagree in sign on the same token, so a single combined scalar advantage produces updates that neither optimize EM nor follow the directive. The separate-loss form in Eq.~\ref{eq:total} sidesteps this by keeping the two signals orthogonal in loss space.

\paragraph{$\lambda$-decay schedules.}
We tested linearly decaying $\lambda_{\mathrm{OPD}}$ from $0.01$ to $0$ over $[0.2T, 0.4T]$ and over $[0.0T, 0.5T]$, on the intuition that OPD should matter most early in training. Both schedules underperformed the constant $\lambda_{\mathrm{OPD}}=0.01$ at val@150; we use the constant schedule in all main results.

\section{Qualitative critique example}
\label{app:example}

Figure~\ref{fig:critique_example} shows a representative failed trajectory from training (HotpotQA) together with the trajectory-level critique returned by the judge. Snippets are abbreviated to fit the column. The hint is then prepended to the student's context for every \texttt{<search>} step in this trajectory, and the OPD loss pushes the student toward the teacher's hint-conditioned token distribution on those search tokens only.

\begin{figure}[h]
\centering
\fbox{\begin{minipage}{0.94\linewidth}
\small
\textbf{Question.} \emph{Who was the director of the 1985 film \textit{Brazil}?}\\
\textbf{Gold answer.} Terry Gilliam.\\[2pt]
\hrule
\vspace{3pt}
\textbf{Trajectory} ($r(\tau)=0$):\\[2pt]
\texttt{Step 1.} \texttt{<search>}\emph{Who was the director of Brazil?}\texttt{</search>}\\
\textit{Retrieved:} Doc 1 ``Cinema of Brazil'' (Glauber Rocha); Doc 2 ``Brazil (country)'' (Fernando Meirelles, \emph{City of God}); Doc 3 ``Brazil (1985 film)'' (production background; no director name in snippet).\\[2pt]
\texttt{Step 2.} \texttt{<search>}\emph{director of Brazil}\texttt{</search>}\\
\textit{Retrieved:} Doc 1--2 ``Roberto Azevêdo'' (WTO Director-General); Doc 3 ``Secretary General of Foreign Affairs of Brazil''.\\[2pt]
\texttt{Step 3.} \texttt{<answer>}\emph{Glauber Rocha}\texttt{</answer>}\\[2pt]
\hrule
\vspace{3pt}
\textbf{Judge critique.}\\
\colorbox{gray!12}{\parbox{0.95\linewidth}{\emph{The agent confused the country ``Brazil'' with the 1985 film of the same name. The first query should have disambiguated to the film (e.g.\ ``director of Brazil 1985 film'' or ``Terry Gilliam Brazil 1985''); the second query made the ambiguity worse by retrieving political-figure pages instead.}}}
\end{minipage}}
\caption{Representative failed trajectory and trajectory-level critique. The same critique is broadcast as a hint to all three \texttt{<search>} steps and used to condition the frozen teacher in the OPD loss.}
\label{fig:critique_example}
\end{figure}

\section{Judge prompt and model identifiers}
\label{app:judge}

\paragraph{Judge model identifiers.} For full reproducibility, the exact judge checkpoints used in this work are:
\begin{itemize}[leftmargin=*,itemsep=1pt,topsep=2pt]
\item \textbf{Qwen3.6-27B} (main): \texttt{Qwen/Qwen3.6-27B} on Hugging Face, served via vLLM with \texttt{enable\_thinking=false}.
\item \textbf{Qwen2.5-7B-Instruct} (sensitivity ablation): \texttt{Qwen/Qwen2.5-7B-Instruct}, served via vLLM.
\item \textbf{DeepSeek-V4-Flash} (sensitivity ablation): public DeepSeek API, model identifier \texttt{deepseek-v4-flash}, endpoint \texttt{https://api.deepseek.com/chat/completions}.
\end{itemize}
All judge calls use \texttt{temperature=0.2}, \texttt{max\_tokens=500}, and the prompt below.

\paragraph{Trajectory-judge prompt.} The frozen judge $\pi_J$ is queried once per failed trajectory with the following template (the first and last two search steps are kept verbatim; intermediate steps are abbreviated to control prompt length):

\begin{quote}\small\tt
A search agent failed to answer the question correctly. Diagnose the failure.

Question: \{question\}\\
Correct answer: \{ground\_truth\}

Agent's search trajectory:\\
Step 1: query = "..."\\
\hspace*{8pt}retrieved: ...\\
... omitted N middle steps ...\\
Step k: query = "..."\\
\hspace*{8pt}retrieved: ...

In one or two sentences (max 50 words), describe what the agent should have done differently. Be specific to THIS trajectory --- name actual queries, retrieved evidence, or missing information. Output the critique only --- no preamble, no labels, no markdown.
\end{quote}

The judge's response is parsed as plain text, capped at 500 characters by extracting up to the last sentence boundary, and used directly as the hint $h(\tau)$ that conditions the frozen teacher (Eq.~\ref{eq:tctx}).

\section{Training cost vs.\ canonical OPD}
\label{app:cost}
\method{}'s compute overhead is comparable to or lower than canonical on-policy distillation in two respects. First, the auxiliary teacher in canonical OPD is queried once per response token to obtain $\pi_{\mathrm{ref}}$ log-probabilities; \method{} keeps the same per-token teacher forward but restricts the loss to the search-action token mask $M$, which in our trajectories covers roughly $20$--$30\%$ of response tokens, so the OPD gradient computation is correspondingly cheaper than a full-response OPD. Second, the judge call is paid only on failed trajectories ($r(\tau)=0$, typically $60$--$80\%$ of rollouts early in training and declining as the policy improves); this adds about $15\%$ wall-clock per step in our setup---a one-time cost at training time that disappears at inference, since the judge is not in the rollout loop at evaluation. Hint-conditioned distillation methods of the same family pay a comparable overhead.

\end{document}